%% file: main.tex
\documentclass{article}

     \PassOptionsToPackage{numbers, compress}{natbib}

     \usepackage[preprint]{neurips_2021}

\usepackage[utf8]{inputenc} 
\usepackage[T1]{fontenc}    
\usepackage{hyperref}       
\usepackage{url}            
\usepackage{booktabs}       
\usepackage{amsfonts}       
\usepackage{nicefrac}       
\usepackage{microtype}      

\usepackage{amsmath}        
\usepackage{mathtools}
\usepackage{dsfont}

\usepackage{wrapfig}

\DeclareMathOperator{\MMD}{MMD}
\DeclareMathOperator*{\argmax}{argmax}
\DeclareMathOperator*{\argmin}{argmin}
\DeclareMathOperator{\RL}{RL}
\DeclareMathOperator{\IRL}{IRL}
\DeclareMathOperator{\Geom}{Geom}

\DeclareMathOperator*{\CAMERON}{CAMERON}

\newtheorem{proposition}{Proposition}

\usepackage{algorithm}
\usepackage[noend]{algorithmic}
\usepackage{subcaption}
\usepackage{xcolor}

\title{Offline Inverse Reinforcement Learning}

\author{%
  Firas JARBOUI \\
  ANEO \\
  Centre Borelli - ENS Paris-saclay  \\
  \texttt{firasjarboui@gmail.com} \\
  \And
  Vianey PERCHET \\
  Criteo AI Lab \\
  Crest, ENSAE \\
  \texttt{vianney.perchet@normalesup.org} \\
}

\begin{document}
\maketitle
\begin{abstract}
    The objective of offline RL is to learn optimal policies when a fixed exploratory demonstrations data-set is available and sampling additional observations is impossible (typically if this operation is either costly or rises ethical questions).
    In order to solve this problem, off the shelf approaches require a properly defined cost function (or its evaluation on the provided data-set), which are seldom available in practice.
    To circumvent this issue, a reasonable alternative is to query an expert for few optimal demonstrations in addition to the exploratory data-set. The objective is then to learn an optimal policy w.r.t. the expert's latent cost function. 
    Current solutions either solve a behaviour cloning problem (which does not leverage the exploratory data) or a reinforced imitation learning problem (using a fixed cost function that discriminates available exploratory trajectories from expert ones). 
    Inspired by the success of IRL techniques in achieving state of the art imitation performances in online settings, we exploit GAN based data augmentation procedures to construct the first offline IRL algorithm.
    The obtained policies outperformed the aforementioned solutions on multiple OpenAI gym environments.   
\end{abstract}

\input{Source/Introduction}
\input{Source/Problem}
\input{Source/Off-line_RL}
\input{Source/Off-line_IRL}

\input{Source/Experiments}
\input{Source/conclusion}

\newpage
\bibliography{example_paper}
\bibliographystyle{abbrv}

\newpage
\appendix
\input{Appendix/technical_results}
\input{Appendix/experimental_results}

\end{document}

%% file: Source/Introduction.tex
\section{Introduction}
    The motivations behind offline  reinforcement learning come from many real life problems, such as self-driving cars and medical applications, where collecting data online using low quality intermediate policies is costly or unethical \cite{wu2019behavior, kumar2019stabilizing}. 
    In these use cases, the ground truth cost function is usually unavailable \cite{SqIL, TGR, ORIL}.  To circumvent this issue, a reasonable alternative is to query an expert for a few optimal demonstrations; however trying to greedily imitate it, as in Behaviour Cloning (BC), requires unreasonably large and diverse volumes of expert demonstrations to reach near-optimal~performances\cite{gail}.
    
    As a consequence, an alternative solution seems to be cost shaping, i.e., the estimation/construction of appropriate cost function \cite{TGR,ORIL}. Unfortunately, this is highly non trivial and practical solutions must be found, and Inverse Reinforcement Learning (IRL) appears to be a promising approach. Indeed, in the less challenging online setting, it is well known that solving the IRL problem achieves state of the art performances to learn both the expert policy and an explaining cost function \cite{gail}. However, this requires solving iteratively an RL problem given the current cost, and a discrimination problem given the learned policy $\pi$ to distinguish expert trajectories from $\pi$-generated ones. While the former problem have been studied thoroughly in the offline setting \cite{fu2020d4rl, yu2021combo, kumar2020conservative, yu2020mopo, kidambi2020morel}, the literature is lacking offline solutions for the latter. In order to circumvent this issue, recent progresses proposed well principled heuristics to construct a cost function that is used latter on to solve an offline RL problem \cite{TGR, ORIL}. For example, inspired by the success of Generative Adversarial Imitation Learning (GAIL) in online IRL problems \cite{gail, airl, EAIRL, RAIRL, sGAIL}, a possibility is to learn an adversarial cost function using the available demonstrations. More precisely, Offline Reinforced Imitation Learning (ORIL) constructs a cost function that discriminates expert and exploratory trajectories under the assumption that the latter might contain expert like transitions by using a Positive-Unlabelled loss function \cite{ORIL}. Similarly, Time Guided Rewards (TGR) constructs a reinforcement signal by solving a discrimination problem under the assumption that early expert transitions are not associated with low costs \cite{TGR}. Despite their reported good performances, they do not provide an offline solution to the IRL problem
    
    We answer these criticisms by constructing an offline solution to the discrimination problem that can be used as a sub-routine when solving offline IRL. The algorithm we derived outperformed existing solutions and achieved state of the art performances on several OpenAI environments.

%% file: Source/Problem.tex
\section{Problem Formulation}
    Let us first recall some classical notations and concepts of Markov Decision Processes (MDP) and RL. An infinite MDP $\mathcal{M}=\{ \mathcal{S}, \mathcal{A}, \mathcal{P}, c, \gamma, p_0 \}$ is defined by:
    \begin{description}
        \item[--] $\mathcal{S}$, a state space (either a compact or a finite subset of $\mathds{R}^d$, where $d\in \mathds{N}$ is the state dimension)
        \item[--] $\mathcal{A}$, a the action space (either a finite or a compact subset of $\mathds{R}^{d'}$, where $d' \in \mathds{N}$ is the action space dimension)
        \item[--] $\mathcal{P}$, a the state transition probability distribution: a continuous mapping from $\mathcal{S} \times \mathcal{A}$ to $\Delta(\mathcal{S})$, where $\Delta(\cdot)$ is the set of probability measures over some set,
        \item[--] $c:\mathcal{S}\times \mathcal{A} \to \mathbb{R}$, a continuous cost function,
        \item[--] $p_0 \in \Delta(\mathcal{S})$, an initial state distribution, and $\gamma \in [0,1]$ is the discount factor.
    \end{description}
    A policy $\pi : \mathcal{S} \to \Delta(\mathcal{A})$ is a mapping defining a probability distribution over the action space for any given state. Fixing the policy, the transition kernel and the initial state distribution generates a unique Markov Chain, associated to a probability distribution over the sequences of states denoted by:
    $\mathbb{P}_\pi^t(s,a|s') = \mathbb{P}(s_t=s, a_t=a|s_0=s', \pi, \mathcal{P})$. 
    Given the cost function $c$, the goal of RL is to optimise the entropy regularised cumulative discounted costs \cite{sac, regMDP} defined as the regularised loss:
    \begin{align*}
        \mathcal{L}(\pi,c) := \mathbb{E}_{p_0,\pi}\big[c+\log\pi\big] & := \int_{s_0} p_0(s_0)  \sum_{t=0}^\infty \int_{s, a} \gamma^t \mathbb{P}_\pi^t(s,a|s_0) \Big[c(s,a) + \log(\pi(a|s))\Big] \\
         & = \int_{s_0} p_0(s_0)  \int_{s,a,s_+,a_+} \hspace{-1.1cm} \rho_\pi^\gamma(s,a|s_0)\Big[c(s,a) + \log(\pi(a|s))\Big] 
    \end{align*}
    where $\rho_\pi^\gamma(s,a|s_0) := \sum_{t=0}^\infty \gamma^t \mathbb{P}_\pi^t(s, a|s_0)$ is the $\gamma$-discounted occupancy measure.
    Given an expert policy $\pi_E$, the objective of Maximum Entropy IRL \cite{MEIRL1, MEIRL2} is to find  a cost function $c$ such that the expert policy $\pi_E$ has a low regularised loss $\mathcal{L}(\pi_E,c)$ while other policies incur a much higher loss. Recent progress \cite{GIRL} in IRL propose to optimise a more robust loss that re-weights future states:
    \begin{align*}
        &\mathcal{L}^\eta(\pi,c) := \mathds{E}^{\eta}_{p_0,\pi}\big[ c+\log\pi \big] = \int_{s_0} \hspace{-0.2cm} p_0(s_0)  \int_{s,a,s_+,a_+} \hspace{-1.1cm} \rho_\pi^\gamma(s,a|s_0)P_\pi^{\eta}(s_{+}, a_{+}|s) \Big[c(s_+,a_+) + \log(\pi(a_+|s_+))\Big]
    \end{align*}
    where $P_\pi^{\eta}(s_{+}, a_{+}|s) := \sum_{t=0}^\infty \eta(t) \mathbb{P}_\pi^t(s_{+}, a_{+}|s)$ is the $\eta$-discounted occupancy measure and $\eta$ is a probability distribution over $\mathbb{N}$ that modulates the weighing of future states. 
    Notice that by setting $\eta$ to a Dirac mass at $0$, the obtained loss is equivalent to the classical regularised discounted cumulative rewards $(i.e.\, \mathcal{L}^{\delta_0}(\pi,c)=\mathcal{L}(\pi,c))$. On the other hand, setting $\eta$ to a geometric distribution $(i.e.\,\eta=\Geom(\delta))$ is equivalent to optimising regularised cumulative discounted Q-function values:
    \begin{align*}
        \mathcal{L}^\delta(\pi,c) := \mathcal{L}^{\Geom(\delta)}(\pi,c) = \mathbb{E}_{p_0,\pi}\big[Q_\pi^\delta+\log\pi\big] \; \text{s.t.} \; Q_\pi^\delta(s,a) = \sum_{t=0}^\infty \int_{s', a'} \hspace{-0.3cm} \delta^t \mathbb{P}_\pi^t(s',a'|s,a)c(s',a').
    \end{align*}
    In this paper we adapt the latter loss function. Formally, the associated RL and IRL problems are: $\RL^\delta(c)  := \argmin_{\pi} \mathcal{L}^\delta(\pi,c)$ \text{and} $\IRL^\delta(\pi_E) := \argmax_c  \min_{\pi} \mathcal{L}^\delta(\pi,c) -  \mathcal{L}^\delta(\pi_E,c)   - \psi(c)$ where $\psi$ is a cost penalisation function defined as:
    \begin{equation*}
        \psi(c) = 
            \left\{
            \begin{array}{cc}
                \mathbb{E}_{p_0,\pi_E}^{\eta} [g(c(s,a))] & \textit{ if } c<0 \\
                +\infty &  \textit{ otherwise }
            \end{array}
            \right .
        \text{s.t.} \; g(x) = 
            \left\{
            \begin{array}{cc}
                  -x - \log(1-e^x) &  \textit{ if } x<0 \\
                  +\infty &  \textit{ otherwise }
            \end{array}
            \right .
    \end{equation*}
    This formulation implicitly boils down to learning a policy $\hat{\pi}$ that minimises the worst-case cost weighted divergence \cite{GIRL} $d_c^{\gamma, \delta}(\hat{\pi}\|\pi_E): \mathcal{S}\mapsto\mathbb{R}$ averaged over the initial state distribution $p_0$, where:
        \begin{align*}
            d_c^{\gamma, \delta}(\hat{\pi}\|\pi_E)(s_0)
            &:= \int_{s,a,s_+,a_+} \hspace{-1.1cm} c(s_+,a_+) \Big[  \rho_{\hat{\pi}}^\gamma(s,a|s_0)\rho_{\hat{\pi}}^\delta(s_{+}, a_{+}|s) -\rho_{\pi_E}^\gamma(s,a|s_0)\rho_{\pi_E}^\delta(s_{+}, a_{+}|s) \Big] \\
            & \;= \int_{s_+,a_+} \hspace{-0.6cm} c(s_+,a_+) \Big[  \mu_{\hat{\pi}}^{\gamma, \delta}(s_{+}, a_{+}|s_0) -\mu_{\pi_E}^{\gamma, \delta}(s_{+}, a_{+}|s_0) \Big] \\
            \text{s.t.} \quad \mu_{\pi}^{\gamma, \delta}(s_+, a_+|s_0) & := \sum_{t,k} \gamma^t \delta^k \mathbb{P}_\pi^{t+k}(s_+, a_+|s_0) = \int_{s,a} \rho_\pi^\gamma(s,a|s_0)\rho_\pi^\delta(s_{+}, a_{+}|s)
        \end{align*}
    In fact, optimising the IRL objective can be seen as a min-max optimisation of the Lagrangian~$L^{\gamma, \delta}$: 
    \begin{align*}
        L^{\gamma, \delta}(\pi,c) & :=  \int_{s_0} \hspace{-0.1cm} p_0(s_0) d_c(\pi\|\pi_E)(s_0) + H(\pi) - H(\pi_E)  - \psi(c) \\
        & \;= \mathcal{L}^\delta(\pi,c) -  \mathcal{L}^\delta(\pi_E,c)   - \psi(c) 
    \end{align*}
    where $H(\pi) := \mathds{E}^{\Geom(\delta)}_{p_0,\pi}[\log\pi] = \int_{s_0} p_0(s_0) \mu_{\pi}^{\gamma, \delta}(s, a|s_0)\log\pi(a|s)$ is the entropy regulariser. 
    For this reason, state of the art approaches  consist in a two-step~procedure: 
    
    \textbf{Solving an RL problem: } Given the cost function $c$, an (approximately) optimal policy $\pi$ is learned using classical model-free RL algorithms:
        \begin{align}
            \pi = \argmin_\pi L^{\gamma, \delta}(\pi,c) = \argmin_\pi \mathcal{L}^\delta(\pi,c) = \RL^\delta(c)
        \label{E:policy_step}
        \end{align}
    \textbf{Solving a discrimination problem: } Given expert and $\pi$-generated trajectories, the cost function is updated to discriminate against states that weren't frequently visited by the expert in the sense of $d_c(\pi\|\pi_E)$ using the following loss (we alleviate notations by using $\mu_\pi$ instead of $\mu_\pi^{\delta, \gamma}$): 
        \begin{align}
            c = \argmax_c L^{\gamma, \delta}(\pi,c) = \argmax_{c \in [0,1]^{\mathcal{S}\times\mathcal{A}}} \mathbb{E}_{s,a \sim \mu_{\pi}}\Big[\log c (s,a)\Big]-\mathbb{E}_{s,a \sim \mu_{E}}\Big[\log (1-c)(s,a)\Big] 
        \label{E:cost_step}
        \end{align}
    Naturally, these steps are to be repeated until convergence. However, both steps of this template scheme assumes the ability to sample $\pi$-generated trajectories. Unfortunately, this contradicts the offline assumption of a fixed replay-buffer (i.e. the inability to query the MDP $\mathcal{M}$ for additional trajectories). 
    In the following sections we discuss how to solve these problems offline.

%% file: Source/Off-line_RL.tex
\section{Offline RL (Reinforcement Learning)}
    In model-free offline RL, the difficulty arises from out-of-distribution transitions (taking an action or visiting a state that does not `appear'  in the replay buffer). We discuss in this section state of the art approaches to solve the RL problems in offline settings.
    
    The objective of actor-critic value based RL algorithms  is to iteratively learn the current state-action values function $Q_{\pi_c}^\gamma$ (given the current policy $\pi_c$) and then improve the behaviour by adapting the new policy $\pi_n(.|s) = \delta \big(\argmax_{a} Q_{\pi_c}^\gamma(s,a)\big)$. The policy improvement theorem \cite{sutton} guarantees that this scheme converges to a policy that minimises the value function for all possible state-action pairs, thus minimising $\mathcal{L}^\eta(\pi,c)$ for any distribution $\eta$. In order to learn the value function, the standard approach is to exploit the fact that $Q_\pi^\gamma$ is the unique fixed point of the regularised bellman operator~$\mathcal{B}^\pi$~\cite{regMDP}: 
    \begin{align*}
        & \mathcal{B}^\pi: Q\in\mathbb{R}^{\mathcal{S}\times\mathcal{A}} \mapsto \mathcal{B}^\pi(Q) \in \mathbb{R}^{\mathcal{S}\times\mathcal{A}} \\
        \text{s.t.} &
        \Big[\mathcal{B}^\pi(Q)\Big](s,a) := c(s,a) + \gamma \mathbb{E}_{\substack{a'\sim\pi(.|s')\\s'\sim\mathcal{P}(.|s,a)}}[Q(s',a') + \log\pi(a'|s')]
    \end{align*}
    In practice, state of the art approaches such as Soft Actor Critic \cite{sac}, use a sample based approximation of the regularised Bellman operator $\hat{\mathcal{B}}^\pi$ to update $Q(s,a)$ and update the policy $\pi$ to minimise the learned Q-function. This entails a two-step procedure that solves the RL problem: 
    \begin{description}
        \item[Policy evaluation:] $Q=\argmin_Q \mathbb{E}_{s,a,s'\sim\mathcal{D}}\Big[ \big(Q(s,a) - \hat{\mathcal{B}}^\pi Q(s,a)\big)^2 \Big] $ 
        \item[Policy improvement:] $\pi = \argmin_\pi \mathbb{E}_{s\sim\mathcal{D}, a\sim\pi(.|s)}\Big[ Q(s,a) \Big]$
    \end{description}
    The sample based approximation is defined as $\big[\hat{\mathcal{B}}^\pi(Q)\big](s,a)=c(s,a)+\gamma \big(Q(s',a') + \log\pi(a'|s')\big)$, where the state $s'$ is sampled according to the dynamics and the action $a'$ is sampled according to the policy. Notice that in the policy improvement step, $\pi$ is trained to minimise the current approximation of the Q-function. This process naturally yields out-of-distribution actions for which $Q_\pi$ is under-estimated \cite{levine2020offline, yu2021combo}. In the classical setting, this is not problematic as additional $\pi$-generated trajectories (where the agent visits these state-action pairs) are added to the replay buffer periodically to provide a feed-back loop that rectifies these under-estimations. However in the offline setting where a fixed replay-buffer is available from the beginning, these errors build up to produce a bad approximation of the Q-function, which in turn breaks the policy improvement theorem guarantees. 
    
    In order to avoid this pitfall, previous contributions in offline reinforcement learning developed tools to \textbf{either constraint the learned policy} (by penalising distributions diverging from the exploratory policy --used to generate the replay buffer-- with either explicit f-divergence constraint \cite{jaques2019way, kumar2019stabilizing} or implicit constraints \cite{nair2020accelerating, peters2010relative, wu2019behavior}) \textbf{or learn a conservative Q-function} (by penalising the reward function \cite{kidambi2020morel, yu2020mopo} or the value updates \cite{kumar2020conservative, yu2021combo} in unobserved state action pairs)
    
    Conservative Offline Model-Based Policy Optimisation (COMBO) \cite{yu2021combo} falls in the later category. We propose to use it to alleviate the first issue of offline IRL (approximately solving the RL problem in the offline setting). Our choice is motivated by the theoretical guarantees of safe policy improvement (the learned policy is better than the exploratory policy \cite[Theorem 4]{yu2021combo}) and the tightness of the learned Q-value (the learned Q-function is a tight lower bound of the real one \cite[Theorem 2]{yu2021combo}) as-well as empirical state of the art performances on benchmark data-sets for offline RL \cite{fu2020d4rl, yu2021combo}.
    
    The approach consists of introducing two changes to value based actor-critic RL algorithms:
    
    \paragraph{Approximating the environment dynamics $\mathcal{P}$:} In order to assist the policy search, COMBO approximates the underlying environment dynamics using the offline replay-buffer (with a maximum likelihood estimator $\hat{\mathcal{P}}=\argmax_{\mathcal{P}}\mathbb{E}_{s,a,s'\sim\mathcal{D}}\big[ \mathcal{P}(s'|s,a) \big]$ for instance). The idea is to generate additional k-step roll-outs initialised at randomly sampled states $s\sim\mathcal{D}$ in the learned MDP $\hat{\mathcal{M}}$ (with $\hat{\mathcal{M}}$ being the MDP $\mathcal{M}$ where the dynamics are replaced with $\hat{\mathcal{P}}$ and $k$ is some hyper-parameter). 
    
    \paragraph{Modifying the policy evaluation step:} In order to learn a tight lower bound of the true value function $Q_\pi$, the Q-values of out-of-distribution (OOD) state-action pairs are penalised, while Q-values of at in-distribution pairs are enhanced. Formally, this boils down to the updates provided in Equation \eqref{COMBO_update}. Compared to the classical Bellman updates, three changes are to be noticed: 
    
        \textbf{1--} the Bellman updates are performed for samples drawn according to $d_f=f\rho^\gamma(s,a) + (1-f)\hat{\rho}_\pi^\gamma$ where $\rho^\gamma$ is the occupancy measure of state action pairs from $\mathcal{D}$ and $\hat{\rho}_\pi^\gamma$ is the occupancy measure of state action pairs from the rollouts in the learned MDP $\hat{\mathcal{M}}$, and $f\in[0,1]$;
        
        \textbf{2--} the Q-value of transitions from the roll-outs in $\hat{\mathcal{M}}$ are pushed up (penalising OOD state actions); 
        
        \textbf{3--} the Q-value of transitions from the true MDP ($\mathcal{D}$) are pushed down to balance the penalisation.
        
    \begin{align}
        Q=\argmin_Q \mathbb{E}_{s,a,s'\sim d_f}\Big[ \big(Q(s,a) - \hat{\mathcal{B}}^\pi Q(s,a)\big)^2 \Big] + \beta \Big( \mathbb{E}_{s,a\sim\mathcal{D}}\Big[ Q(s,a)\Big] - \mathbb{E}_{s,a\sim\hat{\rho}_\pi}\Big[ Q(s,a)\Big] \Big) 
        \label{COMBO_update}
    \end{align}
    We propose to use COMBO as a sub-routine of our proposed offline IRL solution. 

%% file: Source/Off-line_IRL.tex
\section{Offline IRL (Inverse Reinforcement Learning)}
    The classical approach to update the cost function (given its `approximately' optimal policy $\pi$) is to discriminate $\pi$-generated trajectories against the expert ones to maximise the divergence $d_c( \pi\| \pi_E )$~\cite{gail, airl, GIRL}. This is done by training a $[0,1]$ discriminator to minimise the binary cross entropy loss between samples from $\mu_{E}$ and $\mu_{\pi}$, and then by using the learned discriminator as the new cost function. 
    However, this approach is not suitable for the offline setting as it uses $\pi$-generated trajectories as a sampling proxy of $\mu_{\pi}$. In this section we discuss an alternative approaches to sample from $\mu_{\pi}$ without querying the true MDP $\mathcal{M}$ for such trajectories. 
    
    For this purpose, remember that 
    $
    \mu_{\pi}^{\gamma, \delta}(s_+, a_+|s_0) = \int_{s,a} \rho_\pi^\gamma(s,a|s_0)\rho_\pi^\delta(s_{+}, a_{+}|s)
    $
    where $\gamma$ is the MDP discount factor and $\delta$ is the discount factor of the geometric distribution $\eta$. This implies that learning the distribution $\mu_{\pi}^{\gamma, \delta}$ reduces to learning $\rho_\pi^\gamma$ and $\rho_\pi^{\delta}$. 
    In the following, we construct a tractable approach to approximate such distributions.
    
    \paragraph{Learning $\rho_\pi^\gamma$ distributions:}
    We propose to decompose the problem into learning a classifier $C^\gamma_\pi$ of such distributions and then learning a generator \hbox{$G:s\in\mathcal{S}\rightarrow (s_+,a_+)\in\mathcal{S}\times\mathcal{A}$} that maximises $\rho_\pi^\gamma(G(s)|s)$ using a Generative Adversarial Network (GAN).
    
    The described decomposition is derived from the following identity:
    \begin{align*}
        \rho_\pi^\gamma(s_+,a_+|s) = P_{\mathcal{D}}(s_+,a_+)\frac{C^\gamma_\pi(s_+,a_+|s)}{1-C^\gamma_\pi(s_+,a_+|s)} \; \text{s.t.} \; C^\gamma_\pi(s_+,a_+|s) = \frac{\rho_\pi^\gamma(s_+,a_+|s)}{\rho_\pi^\gamma(s_+,a_+|s) + P_{\mathcal{D}}(s_+,a_+)} 
    \end{align*}
    where $P_{\mathcal{D}}$ is the state-action distribution in the replay-buffer $\mathcal{D}$.
    Given an off-policy data set, a tractable approach to learn $C^\gamma_\pi$ is to optimise the following importance weighted loss~\cite{Clearning}:  
    \begin{align*}
    \begin{split}
        \mathcal{L}_\pi^\gamma(C) = & \mathbb{E}\Big[ \pi(a_t|s_t) \big[ (1-\gamma) \log C(s_{t+1},a_{t+1}^\pi|s_t) \\
        & + \lfloor w \rfloor \log C(s_+,a_+|s_t) +  \log (1-C(s_+,a_+|s_t)) \big] \Big]
    \end{split}
    & \text{s.t.} \quad w = & \frac{C(s_+,a_+|s_{t+1})}{1-C(s_+,a_+|s_{t+1})} 
    \end{align*}
    where the expectation is taken with respect to $P_\mathcal{D}$ for both $(s_t,a_t)$ and $(s_+, a_+)$, with respect to the dynamics $\mathcal{P}$ for $s_{t+1}$, and with respect to $\pi(.|s_{t+1})$ for $a_{t+1}^\pi$ .
    The notation $\lfloor w \rfloor$ is solely introduced as a reminder that the gradient of an importance weighted objective does not depend on the gradient of the importance weights.
    Naturally, the optimal classifier satisfies $C_\pi^\gamma = \argmin_{C\in[0,1]^{\mathcal{S}\times\mathcal{A}\times\mathcal{S}}} \mathcal{L}_\pi^\gamma(C) $.
    
    Given the classifier $C_\pi^\gamma$, we propose to solve a game between a discriminator $D: (\mathcal{S}\times\mathcal{A}\times\mathcal{S})\to[0,1]$ and a generator $G$ in order to approximate the distribution $\rho_\pi^\gamma$. The goal of the generator $G$ is to produce future state like samples while the discriminator $D$ aims to identify true samples from generated ones. 
    Consider the following score functions for the game:
    \begin{align*}
        V_\pi^\gamma(D,G) = \mathbb{E}_{\substack{s\sim\mathcal{U}(\mathcal{S}) \\ (s_+,a_+)\sim P_\mathcal{D}}}\Big[ \frac{C_\pi^\gamma(s_+,a_+|s)}{1-C_\pi^\gamma(s_+,a_+|s)} \log(D(s_+,a_+|s)) + \log(1-D(G(s)|s)) \Big] 
    \end{align*}
    where $\mathcal{U}(\mathcal{S})$ is the uniform distribution across the state space. Solving this game approximates~$\rho_\pi^\gamma$:
    \begin{proposition}
        $(\Tilde{D}=\frac{1}{2},\Tilde{G}=\rho_\pi^\gamma)$ is a Nash-equilibrium of the following zero-sum game:
        \begin{align*}
            D^* = \argmin_D V_\pi^\gamma(D,G) \quad ; \quad
            G^* = \argmax_G V_\pi^\gamma(D,G)
        \end{align*}
    \end{proposition}
    
    From  early empirical results, it appears that separately training the classifier and using it to train the discriminator/generator pair is unstable, especially when the target distribution $\rho_\pi^\gamma$ has a high variance. Another problem with this approach in practice is that it require training the classifier $C_\pi^\gamma$ up to convergence before optimising the GAN objective $V_\pi^\gamma$. Solving this problem as a sub-routine in each step of an offline IRL algorithm will inevitably incur a high computational cost. 
    In order to solve these problems, we propose to learn a single evaluation function $E$ that will play the roles of both the classifier and the discriminator. Intuitively, this function should be good at distinguishing samples from $\rho_\pi^\gamma$ and samples from $\mathcal{D}$ (which we translate into optimising the loss $\mathcal{L}_\pi^\gamma$) while being able to discriminate against samples from the generator (which we translate into a constraint over the score function $V_\pi^\gamma$). This boils down into the following optimisation problem: 
    \begin{equation*}
    \left\{
        \begin{array}{ll}
            \text{minimise w.r.t. } E: & \mathcal{L}_\pi^\gamma(E) \\
            \text{subject to }: & \mathbb{E}\Big[  w_g \log(E(s_+,a_+|s)) + \log(1-E(G(s)|s)) \Big] < \tau 
        \end{array}
    \right. 
    \end{equation*}
    where $\tau$ is the constraint threshold and $w_g = \frac{E(s_+,a_+|s)}{1-E(s_+,a_+|s)}$. The random variables in the constraint are sampled as described in the score function $V_\pi^\gamma$.
    Using a Lagrange multiplier $\lambda$, the previous optimisation problem is reduced into a global objective function $\mathcal{O}_\pi^\gamma(E,G)$ (with the notation $\lfloor . \rfloor$ as a reminder that the gradient of importance weighted objective does not depend on their gradient):
    \begin{align*}
        \mathcal{O}_\pi^\gamma(E,G) = \mathcal{L}_\pi^\gamma(E) + \lambda \mathbb{E}\Big[ \lfloor w_g \rfloor \log(E(s_+,a_+|s)) + \log(1-E(G(s)|s)) \Big]
    \end{align*}
    This objective function is used to solve a game between the evaluation function $E$ and the generator~$G$:
    \begin{align}
            E^* = \argmin_E \mathcal{O}_\pi^\gamma(E,G) \quad ; \quad
            G^* = \argmax_G \mathcal{O}_\pi^\gamma(E,G)
        \end{align}
        
    \begin{algorithm}
        \caption{Off-policy Idle (Learning the $\gamma$-discounted occupancy measure $\rho_\pi^\gamma$)}\label{Idle}
        \begin{algorithmic}[1]
            \STATE {\bfseries Input:} Trajectories $\tau$, a discount $\gamma$, a policy $\pi$, and a Lagrange multiplier $\lambda$
            \STATE Initialise an evaluation function $E_{\zeta_0}$ and a generator $G_{\nu_0}$
            \FOR{$i \in [1, N]$}
                \STATE Sample transitions $(S_t,A_t,S_{t+1})\sim \tau$, and future state action pairs $(S_+,A_+)\sim\tau$
                \STATE Sample $(t+1)$ actions $A_{t+1}\sim \pi(.|S_{t+1})$
                \STATE Compute the classification weights $W_c = \textsc{stop-gradient}\Big[ \frac{E_{\zeta_{i-1}}(S_+,A_+|S_{t+1})}{1-E_{\zeta_{i-1}}(S_+,A_+|S_{t+1})} \Big]$
                \STATE Sample $(S^+,A^+)\sim \tau$, $S\sim\tau$ and $(S^+_G,A^+_G)\sim G_{\nu_{i-1}}(S)$
                \STATE Compute the discrimination weights $W_d = \textsc{stop-gradient}\Big[ \frac{E_{\zeta_{i-1}}(S^+,A^+|S)}{1-E_{\zeta_{i-1}}(S^+,A^+|S)} \Big]$
                \STATE Update the evaluation parameter $\zeta_i$ to minimise: 
                \begin{align*}
                    \sum \; & (1-\gamma) \log E_{\zeta_i}(s_{t+1},a_{t+1}|s_t) + w_c \log E_{\zeta_i}(s_+,a_+|s_t) +  \log (1-E_{\zeta_i}(s_+,a_+|s_t)) \\ 
                    & + \lambda \big( w_d \log(E_{\zeta_i}(s^+,a^+|s)) + 
                    \log(1-E_{\zeta_i}(s_G^+,a_G^+|s)) \big)
                \end{align*}
                \STATE Update the generator parameter $\nu_i$ to maximise: 
                    $ \sum_{\substack{S}} \log(E_{\zeta_i}(G_{\nu_i}(s)|s))$
            \ENDFOR
            \STATE {\bfseries Return:} $(E_{\zeta_N}, G_{\nu_N})$
        \end{algorithmic}
    \label{A:off_policy_Idle}
    \end{algorithm}
        
    \paragraph{Approximate sampling from $\mu_\pi^{\gamma, \delta}$:} Recall that in the offline settings, the difficulty when solving the discrimination problem from Equation \eqref{E:cost_step} is to evaluate $\mathbb{E}_{s,a\sim\mu_\pi}[\log c(s,a)]$ without having access to $\pi$-generated trajectories. To circumvent this issue, we proposed to learn the distributions $\rho_\pi^\gamma$ and $\rho_\pi^\delta$ using the fixed data set of trajectories $\mathcal{D}$: Algorithm \ref{A:off_policy_Idle} provides a tractable approach to approximate these distributions by learning intermediate evaluation function. We denote in the following $(\Bar{E}_\pi^\gamma, \Bar{\rho}_\pi^\gamma)$ (and respectively $(\Bar{E}_\pi^\delta, \Bar{\rho}_\pi^\delta)$) the output of Algorithm \ref{A:off_policy_Idle} given the policy $\pi$, the discount $\gamma$ (respectively the discount $\delta$), and the off-policy trajectories $\mathcal{D}$.
    These components, along with the learned dynamics $\hat{\mathcal{P}}$, can be combined in different ways to approximate sampling $(s_+,a_+)$ from $\mu_\pi^{\gamma, \delta}(.|s_0)$. From our early experiments we identified two efficient and tractable sampling schemes: 
    \begin{description}
        \item[A- Idle approach:] Sample intermediate state action pairs $(s,a)\sim\Bar{\rho}_\pi^\gamma(.|s_0)$ then sample the future state action pairs $(s_+,a_+)\sim\Bar{\rho}_\pi^\delta(.|s)$. We will refer to this approximation as $\mu_{\textit{Idle}}$.
        \item[B- Roll-out approach:] Sample uniformly intermediate state action pairs from the replay buffer $\mathcal{D}$ (i.e. $(s,a)\sim\mathcal{U}(\mathcal{D})$), re-sample them according to the $\gamma$-classifier weights (i.e. $(s,a)\sim\frac{\Bar{E}_\pi^\gamma}{1-\Bar{E}_\pi^\gamma}(s,a|s_0)$), sample $N$ roll-outs of horizon $H$ initialised at $(s,a)$ from the learned MDP $\hat{\mathcal{M}}$ in $\Bar{\mathcal{D}}=\{(s_i^t,a_i^t)\}_{i\leq N, t\leq H}$, and then sample the future states $(s_+,a_+)$ from $\Bar{\mathcal{D}}$ with a uniform distribution over the roll-out indices $i$ and a truncated geometric of parameter $\delta$ over the time indices $t$. We will refer to this approximation as $\mu_{\textit{Roll-out}}$.
    \end{description}
    
    Due to the identity $ \mu_{\pi}^{\gamma, \delta}(s_+, a_+|s_0) = \int_{s,a} \rho_\pi^\gamma(s,a|s_0)\rho_\pi^\delta(s_{+}, a_{+}|s) $, both proposed schemes are valid approximation of $\mu_\pi^{\gamma, \delta}$. In the following we denote by $\hat{\mu}_\pi^{\gamma, \delta}$ any mixture of $\mu_{\textit{Idle}}$ and $\mu_{\textit{Roll-out}}$.
    
    \subsection{CAMERON: Conservative Adversarial Maximum-Entropy inverse Reinforcement learning in Offline settings with Neural network approximators} 
        Now that we introduced the building blocks of offline inverse reinforcement learning, we can combine them  and finish constructing the complete pipeline. Recall that solving the IRL problem is reduced to a min-max optimisation of $L^{\gamma, \delta}(\pi,c)$. In the onlline setting, this is done by iteratively solving an RL and a cost discrimination problem. In the offline setting, given a fixed trajectories data-set $\mathcal{D}=\mathcal{D}_\beta \cup \mathcal{D}_E$ (where $\mathcal{D}_\beta$ is an exploration data-set, and $\mathcal{D}_E$ is an expert data-set), we propose the following two-step procedure: 
        
        \textbf{Solving an offline RL problem: } Given the cost function $c$, an (approximately) optimal policy $\hat{\pi}$ is learned using offline value-based \footnote{Policies that satisfy the Bellman optimality criterion are necessarily optimal in the sens of $\mathcal{L}^\delta(\pi, c)$ \cite{GIRL}} RL algorithms:
            \begin{align}
                \begin{split}
                    \hat{Q} = \argmin_Q \; & \mathbb{E}_{s,a,s'\sim d_f}\Big[ \big(Q(s,a) - \hat{\mathcal{B}}^{\pi_k} Q(s,a)\big)^2 \Big]  + \beta \Big( \mathbb{E}_{s,a\sim\mathcal{D}}\Big[ Q(s,a)\Big] - \mathbb{E}_{s,a\sim\hat{\rho}_{\pi_k}}\Big[ Q(s,a)\Big] \Big) \\
                    \hat{\pi} = \argmin_\pi \; & \mathbb{E}_{s\sim\mathcal{D}, a\sim\pi(.|s)}\Big[ \hat{Q}(s,a) \Big]
                \end{split}
            \label{offline_RL_step}
            \end{align}
        Compared to the solution $\pi$ of Equation \eqref{E:policy_step}, the obtained policy $\hat{\pi}$ inherits the guarantees of conservative offline RL (i.e. the learned Q-function is a tight lower bound of $Q^{\pi}$ and $\hat{\pi}$ is a guaranteed to be better than the exploratory policy used to generate $\mathcal{D}$ at optimising the cost $c$).
            
        \textbf{Solving an offline discrimination problem: } Given expert trajectories and a policy $\pi$, the cost function is updated to discriminate against $\hat{\mu}_{\pi}^{\gamma, \delta}$-generated state action pairs that weren't frequently visited by the expert using the following loss:
        \begin{align}
            \hat{c} = \argmax_c \Hat{L}^{\gamma, \delta}(\pi,c) = \argmax_{c \in [0,1]^{\mathcal{S}\times\mathcal{A}}} \mathbb{E}_{s,a \sim \hat{\mu}_\pi}\Big[\log c (s,a)\Big]-\mathbb{E}_{s,a \sim \mu_{\pi_E}}\Big[\log (1-c)(s,a)\Big]  
            \label{offline_discriminator_step}
        \end{align}
        where the difference between $\hat{L}^{\gamma, \delta}$ and the Lagrangian $L^{\gamma, \delta}$ is that the distribution $\mu_\pi$ is replaced with it's approximation $\hat{\mu}_\pi$. Intuitively, the gap between solving Equation~\ref{offline_discriminator_step} instead of Equation~\ref{E:cost_step} is as tight as the concentration of the learned distribution $\hat{\mu}_\pi^{\gamma, \delta}$ around $\mu_\pi^{\gamma, \delta}$. 
        
        In order to stabilise the learning process in the online setting, IRL algorithms \cite{gail, airl, EAIRL, sGAIL} store the sampled trajectories of the learned policies over training in a replay buffer and sample future states from the collected demonstrations when updating the cost function. In our proposed offline IRL solution, we adapt a similar procedure. 
        At each iteration,  approximate future state samples of $\mu_\pi^{\gamma, \delta}$ (where $\pi$ is either the current policy or an exploration policy used to generate $\mathcal{D}_\beta$) are stored in a cost replay buffer $\mathcal{D}_{\textit{cost}}$ using: \textbf{-1} the exploratory data set with probability $f_{\mathcal{D}}$; \textbf{2-} the Idle approximation $\mu_{\textit{Idle}}$ with probability $f_A$; and \textbf{-3} the roll out approximation $\mu_{\textit{Roll-out}}$ with probability $f_B$. Naturally, these probabilities sum up to one: $f_{\mathcal{D}}+f_A+f_B = 1$. Afterwards, the costs are updated as follows:
        \begin{align}
            \hat{c} = \argmax_{c \in [0,1]^{\mathcal{S}\times\mathcal{A}}} \mathbb{E}_{s,a \sim \mathcal{D}_{\textit{cost}}}\Big[\log c (s,a)\Big]-\mathbb{E}_{s,a \sim \mu_{\pi_E}}\Big[\log (1-c)(s,a)\Big]  
            \label{offline_discriminator_step_CAMERON}
        \end{align}
        
        Early empirical investigations indicate  that diversifying the sources of future state samples when filling the cost replay buffer lead to an improved performance. For this reason, we advise using a balanced mixture of the proposed approximations (i.e. $f_{\mathcal{D}}=f_A=f_B=\frac{1}{3}$). This is further confirmed in the experimental section.
        
        
        In Algorithm \ref{CAMERON}, we provide a pseudo-code to summarise the workflow of the proposed solution. 
        
        \begin{algorithm}
            \caption{$\CAMERON$}\label{CAMERON}
            \begin{algorithmic}[1]
                \STATE {\bfseries Input:} Expert data-set $\mathcal{D}_E$, Exploratory data-set $\mathcal{D}_\beta$, initial policy $\pi_{\theta_0}$, initial discriminator function $D_{w_0}$, initial classifiers $(C^\gamma, C^\delta)$, initial generator $(\rho^\gamma, \rho^\delta)$, and probabilities $f_{\mathcal{D}}, f_A, f_B$: 
                \STATE $\mathcal{D}\leftarrow \mathcal{D}_E \cup \mathcal{D}_\beta$, initialise an empty roll-out data-set $\hat{\mathcal{D}}$ and an empty replay buffer $\mathcal{D}_{\textit{cost}}$.
                \STATE Learn the dynamics $\hat{\mathcal{P}}$ using MLE and the data-set $\mathcal{D}$
                \FOR{$i \in [1, N]$}
                    \STATE Store roll-outs $\hat{\tau} \sim (\pi_{\theta_i},\hat{\mathcal{P}})$ in  $\hat{\mathcal{D}}$
                    \STATE Update $(C^\gamma, \rho^\gamma)$ with $\textit{Off-policy Idle}(\mathcal{D}, \pi_{\theta_i}, \gamma)$ and $(C^\delta, \rho^\delta)$ with $\textit{Off-policy Idle}(\mathcal{D}, \pi_{\theta_i}, \delta)$ 
                    \STATE Store in $\mathcal{D}_{\textit{cost}}$ future state samples from $\mathcal{D}_\beta$, $\mu_{\textit{Idle}}$ and $\mu_{\textit{Roll-out}}$ according to $f_{\mathcal{D}}$, $f_A$, and $f_B$
                    \STATE Update the policy parameter $\theta_i$ using Equation \eqref{offline_RL_step} 
                    \STATE Update the cost parameter $w_i$ using Equation \eqref{offline_discriminator_step_CAMERON} 
                \ENDFOR
                \STATE {\bfseries Return:} $(\pi_{\theta_N}, D_{w_N})$
            \end{algorithmic}\label{Algo:Megan}
        \end{algorithm}

%% file: Source/Experiments.tex
\section{Experiments}
    In this section, we analyse empirically the performances of $\CAMERON$. In the offline setting, the choice of the data-set has a major impact on the performances. Naturally, given a fixed algorithm, diverse demonstrations yield better performances \cite{yu2021combo, fu2020d4rl, ORIL, TGR}. For this reason, we focus in this section on a fixed set of trajectories and we provide in the appendix ablation analysis on the impact of using alternative ones. 
    Previous contributions in the offline settings constructed data-sets on which they evaluated performances. For the sake of reproducibility, we exploit the openly sourced demonstrations of D4RL \cite{fu2020d4rl} on which a wide range of offline algorithms have been evaluated. In our experimental setup, the expert demonstrations are the same as theirs and the exploratory ones are a concatenation of their medium and random data-sets (respectively generated with a policy trained to the third of the optimal performances and a random policy). We focused in our analysis on three MuJoCo-based openAI environments with varying complexity: \textbf{-1} the Ant environment (with a state action space of dimension $118$), \textbf{-2} the Half-Cheetah environment (with a state action space of dimension $23$), and \textbf{-3} the Hopper environment (with a state action space of dimension $14$).
    
    The following sections aim at answering the following questions:
    \begin{description}
        \item[-1] Can Algorithm \ref{A:off_policy_Idle} (Idle) efficiently approximate $\rho_\pi^\gamma$ using an off-policy data set of demonstrations ?  
        \item[-2] Can Algorithm \ref{CAMERON} ($\CAMERON$) approximate the expert behaviour better than alternative offline imitation learning procedures?
        \item[-3] Does removing one of the sources when storing future states in $\mathcal{D}_{\textit{cost}}$ improves performances ?
    \end{description}
    
    \subsection{The ability of Idle to approximate future state distributions $\rho_\pi^\gamma$}
        In order to solve the discrimination problem offline, we proposed to approximate the distribution $\rho_\pi^\gamma$ using the off-policy Idle procedure as a sub-routine of $\CAMERON$. The efficiency of this approach is tightly linked to the divergence between the learned approximation $\Bar{\rho}_\pi^\gamma$ and the ground truth. 
        
\begin{figure}
        \centering
        \begin{subfigure}[b]{.32\linewidth}
            \centering
            \includegraphics[width=\linewidth]{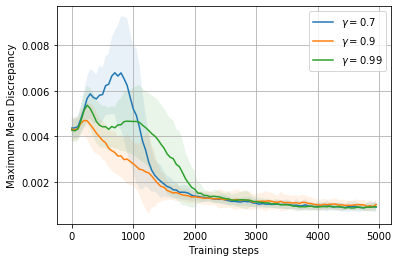}
            \caption{Ant environment}
            \label{idle_perf_ant}
        \end{subfigure}
        \begin{subfigure}[b]{.32\linewidth}
            \centering
            \includegraphics[width=\linewidth]{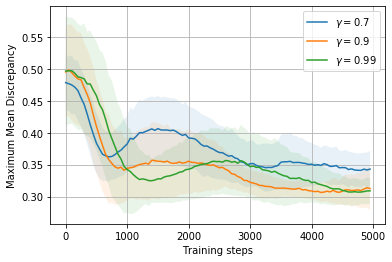}
            \caption{Half-Cheetah environment}
            \label{idle_perf_half_cheetah}
        \end{subfigure}
        \begin{subfigure}[b]{.32\linewidth}
            \centering
            \includegraphics[width=\linewidth]{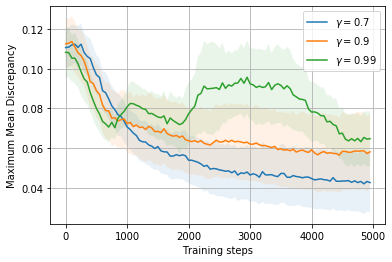}
            \caption{Hopper environment}
            \label{idle_perf_hopper}
        \end{subfigure}
        \caption{ \textbf{MMD divergence as a function of the Idle procedure learning steps:} The obtained approximation effectively reduced the divergence to a level comparable to the ground truth }
        \label{idle_perf}
        \end{figure}

        To evaluate the performances of Algorithm \ref{A:off_policy_Idle}, we computed the Maximum Mean Discrepancy\footnote{A formal reminder on the definition of MMD divergence is provided in the Appendix for completeness.} $\MMD(\rho_\pi^\gamma|\Bar{\rho}_\pi^\gamma)$ over the training iterations. 
        We averaged the results across a set of $20$ policies obtained during the training of an online IRL algorithm.
        For each of these policies, we used the same offline data-set (of expert, medium and random demonstrations) to learn the approximation $\Bar{\rho}_\pi^\gamma$ and we generated additional on-policy demonstrations that we used solely to measure the divergence.
        We conducted this experience for various discount factor values $\gamma$ and reported the results in Figure \ref{idle_perf}.
        
        Agnostic of the discount factor, the learned approximation produced similar future states to those observed in on-policy trajectories (as the measured divergence was reduced by a factor of $2$ to $3$ over the training). In all the environments, approximating high variance $\rho_\pi^\gamma$ distributions (i.e. when the discount $\gamma$ is close to $1$) took the longest. In the considered environments, training the Idle procedure for $5000$ iterations seemed to produce a faithful approximation. 
    
    \subsection{Performance improvement using CAMERON}
        In this section, we compare the performances of $\CAMERON$ with those of current solutions to the offline imitation problem (ORIL \cite{ORIL} and TGR \cite{TGR})\footnote{We provide in the Appendix a reminder of the associated losses for completeness.}. We also evaluate the performances of COMBO (state of the art offline RL) as it provides a performance upper bound in our setting.
        
\begin{figure}
            \centering
            \includegraphics[width=\linewidth]{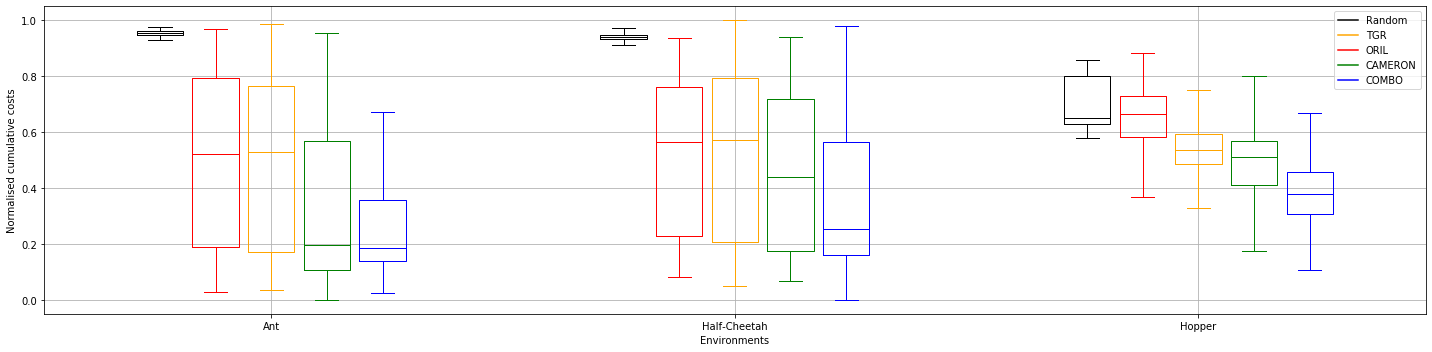}
            \caption{ \textbf{CAMERON outperforms current solutions to the offline imitation learning problem} }
            \label{CAMERON_perf}
        \end{figure}

        For each of the proposed algorithms, we solve the problem using five random seeds and save the best performing policy at each run. These policies are latter on used to generate $200$ trajectories composed of $1000$ transitions. The obtained cumulative costs of these demonstrations ($1000$ per algorithm) are then compiled together to produce the normalised cumulative cost box-plots reported in Figure \ref{CAMERON_perf}. 
        
        In each of the considered environments, CAMERON outperformed the available baselines (ORIL and TGR) and produced competitive performances when compared to COMBO.

    \subsection{Ablation analysis }
        \begin{wrapfigure}[14]{r}{0.45\textwidth}
            \includegraphics[width=0.45\textwidth]{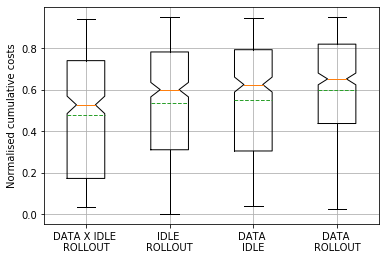}
            \caption{\textbf{Ablation analysis}}
        \label{HC_ablation}
        \end{wrapfigure}
        
        In this section, we analyse the impact of diversifying the sources when filling the cost replay buffer $\mathcal{D}_{\textit{cost}}$. We conduct the same experiment described above on the Half-Cheetah environment using multiple instances of the CAMERON algorithm. In each variation we used a different mixture of $f_{\mathcal{D}}, f_A$ and $f_B$. In Figure \ref{HC_ablation}, we reported the normalised cumulative costs of each configuration. The orange solid lines and the green dashed lines are respectively the median and mean performance, while the notches delimit the $95\%$ confidence interval of the median.  
        
        As expected, each of the proposed approximations (the exploratory data-set $\mathcal{D}_\beta$, the roll-out approximation $\mu_{\textit{Roll-out}}$, and the Idle approximation $\mu_{\textit{Idle}}$) is essential to achieve the best possible performance. Quite interestingly, setting $f_{\mathcal{D}}$ to zero (i.e. not using the available exploration data-set explicitly in the discrimination problem) seems to cause the lowest drop of performance. Intuitively, this entails from the agnostic nature of this source of future samples: even if it does provide relevant observations against which to discriminate, it does not adapt their generations to the current policy. 

%% file: Source/conclusion.tex
\section{Conclusion}
    In this paper we tackled the Inverse Reinforcement Learning problem in the offline setting. 
    The contribution of this work is two folds: we provided a tractable approach to approximate the future state distributions of a given policy using an off-policy data set of demonstrations, and we constructed the first offline IRL algorithm that showcases enhanced performances when compared to alternative offline imitation learning algorithms. An interesting future direction of research consists in enriching the proposed procedures with recent breakthroughs in the online IRL and the generative adversarial networks literature. 

%% file: Appendix/technical_results.tex
\section{Supplementary technical details}

\subsection{Proof of Proposition 1:}
    In order to derive this proposition, notice that:
    \begin{align*}
        \frac{C_\pi^\gamma(s_+,a_+|s)}{1-C_\pi^\gamma(s_+,a_+|s)} & =   \frac{\rho_\pi^\gamma(s_+,a_+|s)}{\rho_\pi^\gamma(s_+,a_+|s) + P_{\mathcal{D}}(s_+,a_+)} 
        \frac{\rho_\pi^\gamma(s_+,a_+|s) + P_{\mathcal{D}}(s_+,a_+)}{P_{\mathcal{D}}(s_+,a_+)} \\
        & = \frac{\rho_\pi^\gamma(s_+,a_+|s)}{P_{\mathcal{D}}(s_+,a_+)}
    \end{align*}
    Which implies that we can re-write $V_\pi^\gamma$ as:
    \begin{align*}
        V_\pi^\gamma(D,G) & = \mathbb{E}_{\substack{s\sim\mathcal{U}(\mathcal{S}) \\ (s_+,a_+)\sim P_\mathcal{D}}}\Big[ \frac{C_\pi^\gamma(s_+,a_+|s)}{1-C_\pi^\gamma(s_+,a_+|s)} \log(D(s_+,a_+|s)) + \log(1-D(G(s)|s)) \Big] \\
        & = \mathbb{E}_{\substack{s\sim\mathcal{U}(\mathcal{S}) \\ (s_+,a_+)\sim P_\mathcal{D}}}\Big[ \frac{\rho_\pi^\gamma(s_+,a_+|s)}{P_{\mathcal{D}}(s_+,a_+)} 
        \log(D(s_+,a_+|s)) + \log(1-D(G(s)|s)) \Big] \\
        & = \mathbb{E}_{\substack{s\sim\mathcal{U}(\mathcal{S}) \\ (s_+,a_+)\sim \rho_\pi^\gamma(s_+,a_+|s)}}\Big[
        \log(D(s_+,a_+|s)) + \log(1-D(G(s)|s)) \Big]
    \end{align*}
    
    The latter formulation is exactly a Conditional Generative Adversarial Network (C-GAN) objective function \cite{CGAN} which verifies the same properties of classical GAN formulation \cite{GAN}. 
    In other words, an oblivious discriminator ($\Tilde{D}=\frac{1}{2}$) and the ground truth distribution ($\Tilde{G}=\rho_\pi^\gamma(s_+,a_+|s)$) are a Nash equilibrium of the zero-sum game valued with $V_\pi^\gamma$.
    
\subsection{Sampling future states using a fixed off-policy data set of demonstrations:}
    Given the currently learned policy $\pi$, solving the discrimination problem (in order to update the cost function) boils down to how well we can approximate future state distribution (according to either $\rho_\pi$ or $\mu_\pi$) using an offline data-set of demonstrations $\mathcal{D}$. 
    Without having access to $\pi$-generated trajectories in the ground truth MDP $\mathcal{M}$, we identified two possible routes for this purpose: 
    \begin{description}
        \item[1-] Generate trajectories using the policy $\pi$ and the learned dynamics $\hat{\mathcal{P}}$ (i.e. in the learned MDP $\hat{\mathcal{M}}$) 
        \item[2-] Approximate the distributions using a GAN-based approach
    \end{description}
    
    \paragraph{Exploiting the learned dynamics:}
    The first approach is limited due to cumulative errors over time. In fact, as the trajectory horizon increases, the obtained trajectories (in $\hat{\mathcal{M}}$) deviates from real ones. 
    
    For this reason, we are only limited to short horizon roll-outs (at most $10$ steps). This means that we can only approximate $\rho_\pi^\gamma$ when the discount $\gamma$ is low (at most $0.9$ in practice) if we want to exploit the learned dynamics in the discrimination problem. 
    For this reason, and to cover the full extent of a desired performance horizon (typically $1000$ steps in the D4RL data sets \cite{fu2020d4rl}) we need to samples probable future observations and perform the short roll-outs from these states: this is the roll-out approach described in the paper (denoted $\mu_{\textit{Roll-out}}$) which naturally approximates the distribution $\mu_\pi$.
    
    We emphasise that the classical IRL formulation (that minimise the divergence in the sense of $\rho_\pi$) does not provide a rational for exploiting such information. On the other hand, the $\eta$-optimality framework provides a rational paradigm in which we can efficiently exploit $\mu_\pi$ when solving the discrimination problem. For example, by setting $\gamma=0.99$ and $\eta=0.9$ (as in all our experiments using CAMERON) we can use the small horizon roll-outs to approximate $\rho_\pi^\eta$ and use the Idle algorithms evaluation function $E_\pi^\gamma$ to samples probable future states. 
    
    \paragraph{Learning the distributions:} The Idle procedure described in the paper, provides a tractable and efficient approach to approximate $\rho_\pi^\gamma$ for any discount value. Recall that sampling future states according to $\mu_\pi$ is equivalent to sampling an intermediate state $s$ according to $\rho_\pi^\gamma(.|s_0)$ and then sampling the future state according  to $\rho_\pi^\eta(.|s)$. This can be done by exploiting directly the learned generators as proposed in the Idle approach proposed in the paper (denoted $\mu_{\textit{Idle}}$). Notice that whenever we sample according to $\rho_\pi^\gamma$, we can equivalently sample states uniformly and then re-sample them according to the weights of the evaluation function $E_\pi^\gamma$. However, this requires a larger batch size to achieve similar performances which induces higher computational costs without providing additional performance gain. For this reason we only focus on the fully Idle approach.

\subsection{Discriminator loss in alternative offline imitation algorithms}
    In this section, we recall the formalism of current offline imitation learning solutions. We considered in the experimental section both the ORIL and TGR algorithms as a baseline when we evaluated the performances of CAMERON as they already proved their superiority to the classical Behaviour Cloning (BC) algorithm. In the following we provide the loss function they use to learn the cost function using a data set of expert demonstration and a data set of exploratory ones. Both approaches exploit the obtained cost to learn the optimal policy in an offline fashion. 
    
    \paragraph{ORIL: positive-unlabelled classification loss}
    In ORIL \cite{ORIL} the assumption is that exploration trajectories ($\mathcal{D}_\beta$) might contain expert like transitions. In order to encode this idea into the discrimination loss, the desired cost function is defined as a binary decision problem that distinguish success from failure. The associated loss function is then defined as: 
    \begin{align*}
        \phi \mathbb{E}_{s,a \sim \mathcal{D}_{\textit{success}}}\Big[\log c (s,a)\Big]- (1-\phi)\mathbb{E}_{s,a \sim \mathcal{D}_{\textit{failure}}}\Big[\log (1-c)(s,a)\Big]  
    \end{align*}
    where $\phi$ is the proportion of the trajectory space corresponding to success. The main idea from positive-unlabelled (PU)-learning \cite{bao2018convex, bekker2020learning} that ORIL exploits, is that expectations with respect to the failure data set can be re-written with respect to the success data and unlabelled data as follows:  
    \begin{align*}
        &(1-\phi)\mathbb{E}_{s,a \sim \mathcal{D}_{\textit{failure}}}\Big[\log (1-c)(s,a)\Big] = \\
        & \qquad \qquad = \mathbb{E}_{s,a \sim \mathcal{D}_{\textit{success}}}\Big[\log (1-c)(s,a)\Big] - \phi \mathbb{E}_{s,a \sim \mathcal{D}_{\textit{unlabelled}}}\Big[\log (1-c)(s,a)\Big]
    \end{align*}
    In summary, the loss function used to learn the cost function in ORIL is defined as: 
    \begin{align*}
        \phi \mathbb{E}_{s,a \sim \mathcal{D}_{\textit{success}}}\Big[\log c (s,a)\Big] - \mathbb{E}_{s,a \sim \mathcal{D}_{\textit{unlabelled}}}\Big[\log (1-c)(s,a)\Big] + \phi \mathbb{E}_{s,a \sim \mathcal{D}_{\textit{unlabelled}}}\Big[\log (1-c)(s,a)\Big]
    \end{align*}
    In the experiments, we considered the faction as a hyper-parameter that we fixed at $\phi=0.5$ as proposed in ORIL.
    
    \paragraph{TGR: time-guided rewards}
    In TGR \cite{TGR}, the used heuristic is that early expert demonstrations are not necessarly associated with low costs. For this reason the learned cost is expected to assign high costs for both early expert transitions $(t<t_0)$ and exploratory transitions while it assigns low costs for the remaining expert demonstrations $(t\geq t_0)$. The associated loss is the following: 
    \begin{align*}
        \mathbb{E}_{s,a \sim \mathcal{D}_{\textit{expert}}; t\geq t_0}\Big[\log c(s,a)\Big] - \mathbb{E}_{s,a \sim \mathcal{D}_{\textit{exploration}}}\Big[\log (1-c)(s,a)\Big]  - \mathbb{E}_{s,a \sim \mathcal{D}_{\textit{expert}}; t < t_0}\Big[\log (1-c)(s,a)\Big]  
    \end{align*}
    In the experiments, we treated the cutting threshold $t_0$ as a hyper-parameter that we fixed at $t_0=50$ in the considered domains. In the original work, a refinement procedure is also proposed to improve robustness the algorithm: it helped stabilising the obtained returns but did not improve performances. For this reason we only compared our IRL solution to the unrefined version of the loss. 
    
\subsection{Maximum Mean Discrepancy evaluation}
\label{A:MMD}
    Formally, given a  reproducing kernel Hilbert space (RKHS) of real-valued functions $\mathcal{H}$, the MMD between two distributions $P$ and $Q$ is defined as: $\MMD_{\mathcal{H}}(P, Q) = \sup_{f\in\mathcal{H}} \mathbb{E}_{X\sim P}[f(X)] - \mathbb{E}_{Y\sim Q}[f(Y)]$. Recall that the reproducing property of RKHS, implies that there is a one to one correspondence between positive definite kernels~$k$ and RKHSs $\mathcal{H}$ such that every function $f\in\mathcal{H}$ verifies $f(x)=\langle f,k(.,x)\rangle_{\mathcal{H}}$ (where $\langle\,,\rangle_{\mathcal{H}}$ denotes the RKHS inner product). We propose to evaluate the MMD using a kernel two-sample test with the following unbiased estimator~\cite{MMD}:
    \begin{align*}
        \MMD_{\mathcal{H}}^2(P, Q) = \frac{1}{N(N-1)} \sum_{i\neq j} k(x_i, x_j) +  \frac{1}{N(N-1)} \sum_{i\neq j} k(y_i, y_j) - \frac{1}{N^2} \sum_{i, j} k(x_i, y_j)
    \end{align*}
    where $(x_i)_{i=0}^N$ are sampled according to $P$ and $(y_i)_{i=0}^N$ are sampled according to $Q$.
    In the experimental analysis, we only consider the RKHS associated with the radial basis function $k(x,y) = \exp(\|x-y\|^2/d)$ (where $d$ is the dimension of the variables $x$ and $y$). 

%% file: Appendix/experimental_results.tex
\section{Supplementary experimental details}
\subsection{Hyper-parameters}
    In this section we provide implementation details as well as the used hyper-parameters. 
    
    Expert demonstrations and exploratory ones (medium and/or random) from D4RL \cite{fu2020d4rl} are stored in separate replay buffers. Both the actors and critics of COMBO \cite{yu2021combo} are approximated using $3$-layer deep, $64$-neuron wide MLP. The cost functions are approximated using $2$-layer deep, $32$-neuron wide network. The evaluation function and generators of the Idle procedures are represented using $3$-layer deep, $64$-neuron wide feed forward architecture with a Gaussian multi-variate output layer. We use a learning rate of $1e-4$ in all the updates. In all the experiments, we fixed the discount factors at $\gamma=0.99$ and $\eta=0.9$.
    
    At each iteration, the algorithm executes the following steps:
    \begin{description}
        \item[1-] Perform $1000$ update steps of the Idle procedure with a Lagrange coefficient $\lambda=0.03$ and a batch size of $256$. This is done for both discount factors $\gamma$ and $\eta$.
        \item[2-] Sample $128$ roll-outs of horizon $h=5$ initialised at randomly selected states from the offline data and store them in a COMBO replay buffer $\mathcal{D}_{\textit{COMBO}}$.
        \item[3-] Sample $128$ roll-outs of horizon $h=5$ initialised at states sampled according $E_\pi^\gamma$. Store these roll-outs in $\mathcal{D}_{\textit{roll-outs}}$
        \item[4-] Sample $4096$ transition according to $\mu_{\textit{Idle}}$ and store them in $\mathcal{D}_{\textit{Idle}}$
        \item[5-] Perform $25$ updates for the discriminator with a batch size $512$. Each policy sample is a mixture of $f_A$ randomly selected transition from $\mathcal{D}_{\textit{Idle}}$, $f_B$ randomly selected transition from $\mathcal{D}_{\textit{roll-outs}}$ and $f_{\mathcal{D}}$ transition sampled according to $\mu_\pi$ from the offline replay. The expert samples exploit the expert offline data set to approximate $\mu_{\pi_E}$.
        \item[6-] Perform $500$ updates for the actors and critics using a batch size of $512$ with a conservatism coefficient $\beta=5$ for both the ant and hopper environment, and $\beta=1$ for the half-cheetah environment. The samples are from the offline data-sets and $\mathcal{D}_{\textit{COMBO}}$.
    \end{description}
    In all environments, we run the algorithm for $1500$ iteration and we saved the best performing policy in the learning process by evaluating the average returns of each policy in the true environment on $10$ trajectories at each step. 
    
\subsection{Learning the dynamics}
\label{sec:roll_out}
    The roll-outs are generated using the learned dynamics $\hat{\mathcal{P}}$. The dynamics are approximated using both the expert and the exploratory demonstrations using $4$-layer deep, $200$-neuron wide MLP by maximising the log-likelihood of the available transitions in the data. Similarly to COMBO, we learned for each data-set (expert + random (and/or) medium demonstrations) $7$ models and only kept $5$ based on validation errors. Whenever we sample a step in the learned MDP $\hat{\mathcal{M}}$, we randomly select one of the $5$ models and produce a new observation.

\subsection{Varying the exploration data-set diversity:}
\begin{figure}
        \centering
        \begin{subfigure}[b]{\linewidth}
        \includegraphics[width=\linewidth]{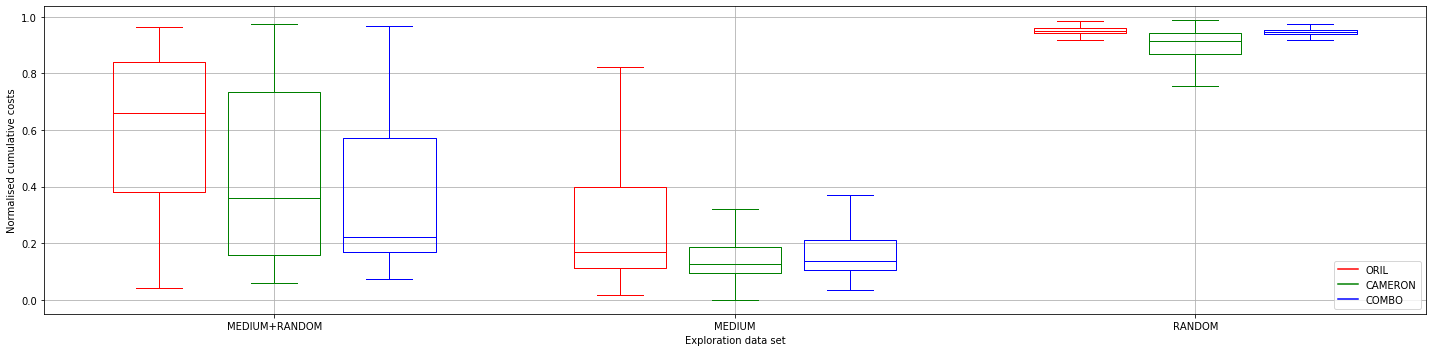}
        \caption{ \textbf{Ant Environment} }
        \label{ablation_ant_perf}
        \end{subfigure}
        
        \begin{subfigure}[b]{\linewidth}
        \includegraphics[width=\linewidth]{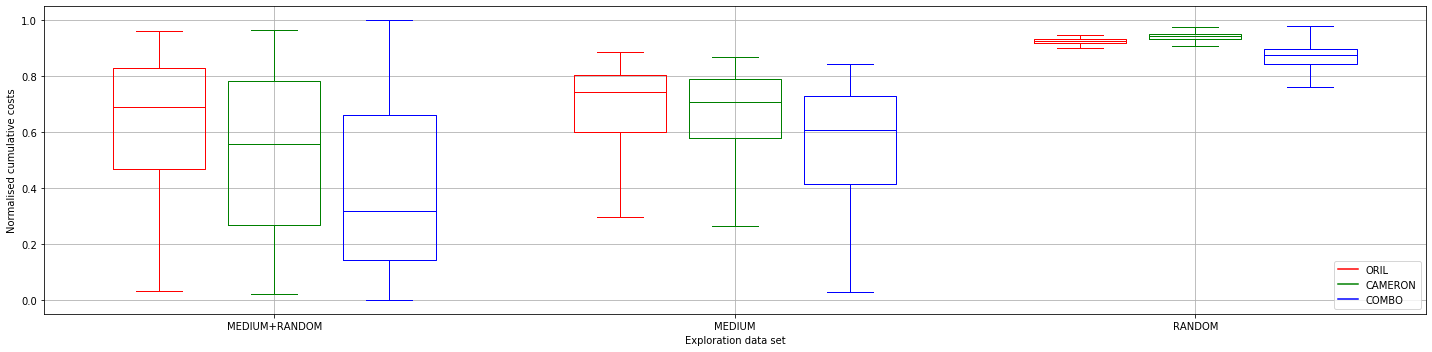}
        \caption{ \textbf{Half Cheetah Environment} }
        \label{ablation_ant_perf}
        \end{subfigure}
        
        \begin{subfigure}[b]{\linewidth}
        \includegraphics[width=\linewidth]{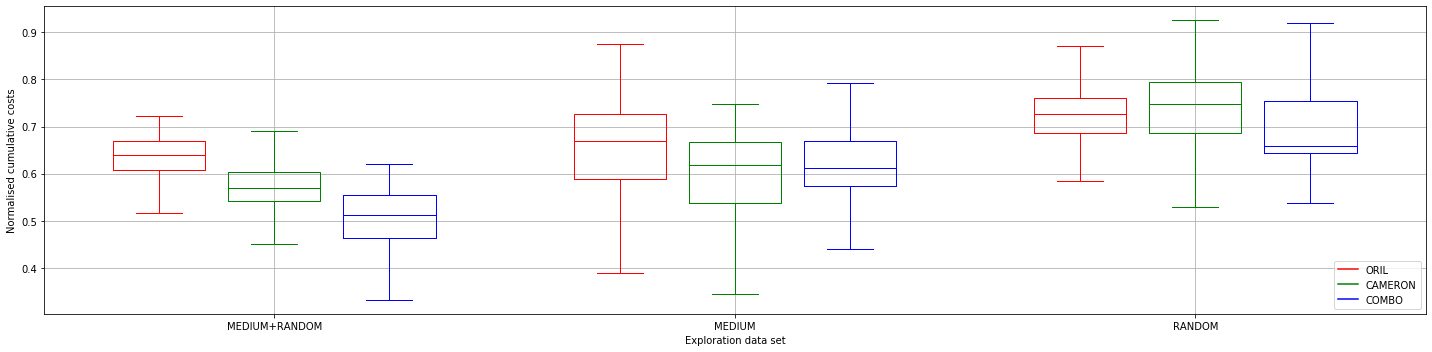}
        \caption{ \textbf{Hopper Environment} }
        \label{ablation_ant_perf}
        \end{subfigure}
    \caption{ \textbf{Varying the diversity of the exploration data set} }
    \label{ablation}
    \end{figure}

    In this section we investigate the impact of varying the diversity of the exploration data set $\mathcal{D}_\beta$. Recall that in the experimental section, we used both the medium and the random trajectories provided in D4RL \cite{fu2020d4rl} to build the exploration data in addition to the expert trajectories. We reproduced the same experiments using each of the random and medium data set individually to construct the exploration replay buffer. For the sake of simplicity we only compare CAMERON to COMBO (offline RL) and ORIL (offline imitation learning). The obtained performances are reported in Figure \ref{ablation}.  
    
    Notably, degrading the diversity of the exploration data is harmful to the performances of offline solutions. Not only this degrades the ability of solving the RL problem (using random trajectories instead of medium ones causes a drastic drop of COMBO performances) but it also entails a worst cost function (as the same observation hold true for ORIL). 
    
    We also notice that agnostic of the used environment, CAMERON outperformed ORIL when we used the medium trajectories. It also achieved competitive performances (if not slightly better in both the ant and hopper environments) to those showcased by COMBO. On the other hand, when we used the random trajectories for the exploration we noticed that ORIL outperformed by a small margin CAMERON (with an exception in the Ant environment). This is explained by the performance drop of the Idle procedure when low quality trajectories are used to approximate future state distributions. 
    
